\documentclass[preprint,12pt,3p]{elsarticle}

\usepackage{epsfig}
\usepackage{subcaption}
\usepackage{calc}
\usepackage{amssymb}
\usepackage{amstext}
\usepackage{amsmath}
\usepackage{amsthm}
\usepackage{multicol}
\usepackage{pslatex}
\usepackage{balance}
\usepackage{float}
\usepackage{siunitx}
\usepackage{placeins}
\usepackage{cleveref}
\usepackage{diagbox}
\usepackage{pdflscape}

\begin{document}

\begin{frontmatter}
\title{Cardiopulmonary resuscitation quality parameters from motion capture data using Differential Evolution fitting of sinusoids}

\author[label1,label2]{Christian Lins\corref{cor1}} 
\address[label1]{Carl von Ossietzky University Oldenburg, Assistance Systems and Medical Device Technology, Dep. Health Services Research, Ammerl\"ander Heerstr. 140, 26129 Oldenburg, Germany}
\address[label1a]{Carl von Ossietzky University Oldenburg, Medical Informatics, Dep. Health Services Research, Ammerl\"ander Heerstr. 140, 26129 Oldenburg, Germany}
\address[label2]{OFFIS - Institute for Information Technology, Div. Health, Escherweg 2, 26121 Oldenburg, Germany\fnref{label4}}
\address[label3]{Nara Institute of Science and Technology, Interactive Media Design Laboratory,\\8916-5, Takayama, Ikoma, Nara, 630-0192, Japan\fnref{label4}}

\cortext[cor1]{Corresponding author}
\fntext[invited]{This paper is an extended, improved version of the paper \emph{Determining Cardiopulmonary Resuscitation Parameters with Differential Evolution Optimization of Sinusoidal Curves} presented at AI4Health 2018 workshop and published in: BIOSTEC 2018, Proceedings of the 11th International Joint Conference on Biomedical Engineering Systems and Technologies, Volume 5: HEALTHINF, Funchal, Madeira, Portugal, 19-21 January, 2018, pp. 665-670, ISBN: 978-989-758-281-3, INSTICC, 2018.}

\ead{christian.lins@offis.de}

\author[label3]{Daniel Eckhoff}

\author[label1a]{Andreas Klausen}

\author[label1]{Sandra Hellmers}

\author[label1,label2]{Andreas Hein}

\author[label1]{Sebastian Fudickar}

\begin{abstract}
Cardiopulmonary resuscitation (CPR) is alongside electrical defibrillation the most crucial countermeasure for sudden cardiac arrest, which affects thousands of individuals every year.
In this paper, we present a novel approach including sinusoid models that use skeletal motion data from an RGB-D (Kinect) sensor and the Differential Evolution (DE) optimization algorithm to dynamically fit sinusoidal curves to derive frequency and depth parameters for cardiopulmonary resuscitation training. It is intended to be part of a robust and easy-to-use feedback system for CPR training, allowing its use for unsupervised training. The accuracy of this DE-based approach is evaluated in comparison with data of 28 participants recorded by a state-of-the-art training mannequin. We optimized the DE algorithm hyperparameters and showed that with these optimized parameters the frequency of the CPR is recognized with a median error of $\pm 2.9$ compressions per minute compared to the reference training mannequin.
\end{abstract}


\end{frontmatter}

\section{Introduction}
\label{sec:introduction}
\noindent
In Europe, sudden cardiac arrest (SCA) is one of the most prominent diseases (350,000-700,000 individuals a year in Europe are affected, depending on the definition \cite{BERDOWSKI20101479, GRASNER2011989, GRASNER2013293}). SCA can significantly affect the independent living of each individual if medical treatment is not available within few minutes \cite{Perkins2015, de2003optimal}. 
With medical treatment including ventilation, medication, and defibrillation (Advanced Life Support, ALS) \cite{soar2015european} there is a high probability of maintaining sufficient blood circulation thus a higher probability of surviving the incident. Medical personnel (including paramedics) is trained in ALS but is usually not immediately available if a cardiac arrest occurs in the field. 
The typical median response time of paramedics is about 5-8 min \cite{Perkins2015}, which is often too long to exclude long-term effects if no bystander (i.e. layman) cardiac massage  meanwhile took place. Immediate help is crucial because the functionality of the cells of the nervous system, including the brain, is reduced after 10 seconds (i.e. loss of consciousness). The death of the cells begins after about 3 minutes \cite{Schmidt2011}. Victims are therefore dependent on the help of bystanders. Thus, as the first minutes (golden minutes) after a cardiac arrest are the most important ones (the likelihood of survival decreases with every minute without cardiac massage  \cite{NEJMoa1405796}), it is crucial that bystanders are well trained to provide appropriate help to the victim. 

\begin{figure}[ht]
    \centering
    \includegraphics[width=\textwidth]{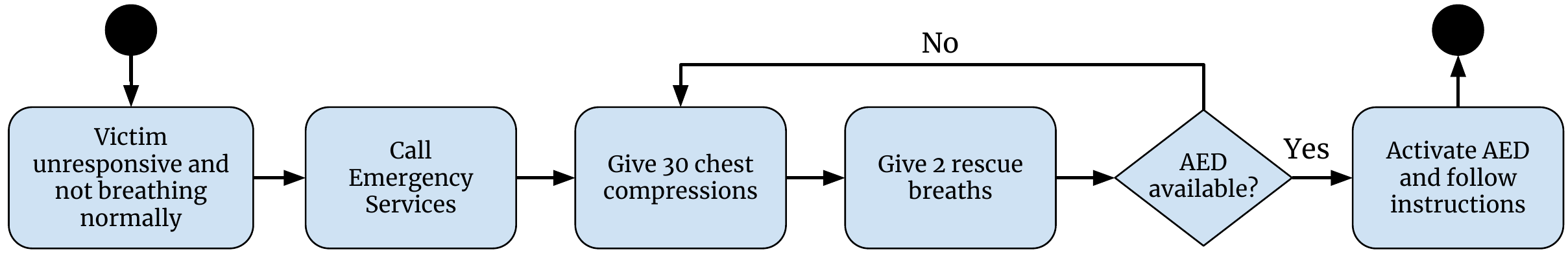}
    \caption{Flow diagram of the Basic Life Support algorithm \cite{Perkins2015}.}
    \label{fig:bls_algo}
\end{figure}

In case of a cardiac arrest, the most critical countermeasure is the \emph{cardiopulmonary resuscitation} (CPR) consisting of cardiac massage (chest compressions) ideally in combination with rescue breathing (Basic Life Support, BLS) \cite{Perkins2015} to sustain a minimal blood circulation to carry oxygen to the nerve cells.
During a cardiac massage, the heart is compressed by orthogonal pressure onto the breastbone. The depth of this compression is ideally \SI{5}{cm} \cite{Perkins2015} to fully eject the blood from the heart. A complete decompression is crucial to fill the heart with blood again. The recommended frequency of the cardiac massage is 100 to 120 compressions per minute (cpm) \cite{Perkins2015}. If a sufficient rescue breathing parallel to the cardiac massage is not possible, i.e. if the victim is not intubated, the thorax compressions will be briefly stopped and will be continued after two accelerated breathings (Figure \ref{fig:bls_algo} summarizes the BLS algorithm).

High-quality CPR improves the outcomes of cardiac arrest, so proper training of medical personnel is as essential as the training of non-specialists, which can offer resuscitation support much faster \cite{NEJMoa1405796}. Since ALS resuscitation training, due to the high material and training costs, is mainly used for medical professionals only, technological training systems might represent a well-suited alternative to train both professionals and non-specialists (CPR bystander). Especially regions without proper training facilities could profit from feasible, low-cost training technologies. 
CPR can be trained with simulation mannequins, which typically provide real-time feedback of the quality of cardiac massage and ventilation. 
The mannequins simulate a patient and can evaluate the training situation solely with their internal sensors. Due to the principle, the training mannequins can make no direct statements about the training behavior of the trainee but only measure its actions indirectly. External measuring, e.g. Motion Capture (MoCap) systems can evaluate the scene and the behavior of the actors, especially for ALS training. However, also for the BLS training, external systems can incorporate additional observations into the quality assessments, such as the posture of the upper body and arms of the trainee in CPR, which has a significant impact on endurance and fatigue of the performer.

An external system could also assist bystander under real conditions with human casualties. A widely used technology is the Automated External Defibrillator (AED), which is used to restore the regular heartbeat in the event of ventricular tachycardia (VT) or ventricular fibrillation (VF) employing an electrical shock. AED complements the ongoing (and still necessary) CPR. For example, an AED device might be equipped with an optical sensor that allows it to capture a large part of the scene and provide audible or visual feedback. Also, old simulation mannequins without an integrated feedback system could continue to be used with an external feedback system.

New approaches with appropriate algorithms are needed to allow MoCap systems to analyze the environment data of scenery and movements during CPR and provide helpful feedback. Such a concept is presented in this paper.

\subsection{Approach}
\noindent
A system with a low-cost RGB-D (RGB + Depth) camera such as the Microsoft Kinect (or even an RGB camera with software skeleton tracking) would provide a suitable alternative to regularly train larger audiences thus improving the overall quality of the CPR within the population. 
As discussed, the CPR training with a system that incorporates external sensors holds advantages over the mannequin training. While mannequins can only provide accurate feedback regarding the compression frequency and depth as acquired by the integrated pressure sensors or IMUs, they cannot detect the relevant space around and at the patient. Therefore, they cannot give feedback on the back and arm position of the rescuer, which has a significant influence on the endurance of the CPR performed and must therefore also be considered. Bystanders performing CPR with an ineffective posture will exhaust rapidly although their CPR frequency and compressions may be appropriate at the beginning. Only an (optical) MoCap system can provide such feedback on the posture. Besides, a motion capture based system may give feedback on the whole process of the ALS in professional training \cite{soar2015european}, e.g. regarding placing of tools, as well as regarding coordination and cooperation with the CPR partner.

The quality parameters for the CPR mentioned above are of different significance.
The most crucial parameters of the CPR are the \emph{frequency} at which the compressions are performed (CCF) and the compression \emph{depth} (CCD) of the chest. Typically, the frequency is given in compressions per minute (cpm) and the compression depth in cm. 
In this paper, we focus on these two parameters and propose a method to derive them from motion data from a RGB-D-based skeleton tracking (Microsoft Kinect v2). 
Thereby, we utilize the periodic nature of the CPR motion and use the time series distances of upper limbs to the floor to fit a sine curve (sinusoidal) model (see Figure \ref{fig:sampleplot}). 

With the proposed approach to monitor CPR executions with external Kinect sensors, CPR quality parameters such as CCF and CCD can be derived without simulation mannequins. Thereby, the discussed disadvantages associated with mannequins are overcome and instead, the system is suitable for CPR training or in-situ observations, e.g. in a shock room.

\subsection{Related Work}
\noindent
RGB-D cameras have already been used in various approaches to support CPR training. Tian et al. for example use Kinect data to model a virtual environment with patient and trainee and drive a haptic device in the real world \cite{tian20143d}. In the virtual environment, the performer can see the CPR on a virtual avatar while responding to the sensation of the haptic device. The authors focus more on the basics of cardiac massage than specific optimization of the training.

Semeraro et al. and Loconsole et al. present results from their system called RELIVE which is similar to our approach \cite{Semeraro2017,loconsole2016relive}. RELIVE uses data from the \mbox{Kinect v1} and extract depth and frequency parameters from the motion data with the intention to improve the quality of CPR (training). In contrast to our approach (usage of the integrated Kinect skeleton tracking), they use the raw RGB pixel and depth image data of the Kinect to identify hands, arms, and the training body. 
The predecessor to RELIVE is probably the Mini-VREM tool, which is a Kinect with software-based audio- and video-feedback-system \cite{Semeraro2013} but requires a marker at the participant's hands (colored bracelet).

Higashi et al. developed and evaluated an augmented reality system that enables the user to correct her or his posture while performing cardiac massage compression \cite{higashi2017}. The focus of this work lays on the correct posture of the performer primarily to differentiate between extended position compression and bent position compression. Unfortunately, they have not yet provided a quantitative evaluation of their system.

Wang et al. propose a real-time feedback system for CPR training, using the \mbox{\emph{Kinect v1}} sensor \cite{wang2017kinect}. The system shows the current compression depth and frequency on a computer screen so that the trainee can adapt her or his actions accordingly. By evaluating the system with 100 healthcare professionals, the authors have shown that the system has significant effects on the CPR quality.
Their system requires a marker on the trainee's hand and the chest of the training mannequin. Additionally, the placement of the Kinect must be carefully chosen here as they directly derive the compression depth from the sensor data. 

Our approach relies on the integrated skeleton tracking of the Kinect (other pose detection methods, e.g. OpenPose \cite{cao2017realtime}, might be feasible as well) and requires no individual markers nor prior calibration. Our method does not directly use the sensor data but fits it to a harmonic sine curve (sinusoid) to make the method robust against sensor noise or false recognition of the skeleton tracking.
However, our approach requires a viable curve fitting algorithm.

To our best knowledge, no one has tried to fit the motion capture (MoCap) data of CPR training to a sinusoid, but there are several possibilities to fit the time series data to a (harmonic) curve.
There are approaches described in signal processing literature to determine the frequency of noisy data. 
Zero crossing is a method that measures the time between two zero crossings of the signal, which is an indication for the frequency of the signal \cite{friedman1994}.
The Fourier decomposition \cite{gauna2016, quinn1997estimation} or a Kalman-Filter \cite{routray2002} has also been investigated. Other methods are a recursive Newton-type algorithm \cite{terzija2003} and adaptive notch filters \cite{dash1998}.
Sachdev and Giray \cite{sachdev1985least} used Taylor-series and least-squares fitting to determine the frequency of a power system. The approach presented in this paper does not require Taylor-series decomposition of a harmonic function but directly fits a sine curve. For this least-squares fitting, a minimization algorithm is needed.
\emph{Differential Evolution} (DE) \cite{Storn1997,price2006} is a heuristic search algorithm, which is well suited for nonlinear functions. It is used for a wide range of applications, e.g. to extract parameters of solar power modules \cite{Ishaque2012}, to find solutions for the equation of Electro Hydraulic Servo Systems \cite{Yousefi2008}, or to fit Bezier curves \cite{Pandunata2010}.

\subsection{Contributions}
\noindent
We show how the Differential Evolution optimization algorithm can be used to fit a sine curve which robustly provides the two wanted parameters. The DE algorithm has to meet some requirements. First, the application must be running in real time to give feedback to the user. This requires a tradeoff between responsiveness and algorithmic runtime. Thus it is necessary to minimize the run time while sustaining sufficient precision. Second, the frequency and depth of the CPR are (in most cases) changing continuously, so the model, i.e. the sinusoid, must be adapted continuously as well.
To summarize our contributions:
\begin{itemize}
    \item We present a practical and widely usable approach to facilitate skeletal motion capture data to derive two quality parameters (compression frequency and compression depth) of the CPR process using a fitted sinusoidal model.
    \item We compared the suitability of wrist, hand, elbow, and shoulder joints to find out which limbs are best for determining CPR quality parameters when using skeletal motion capture. We show the influence of the frequency of model updates $f_U$ and the window size $S_{len}$ on the system accuracy.
    \item We optimized the hyperparameters $NP$ and $G_{Max}$ of the DE algorithm to increase the computational efficiency of our approach.
    \item We evaluated our system approach with data from 28 participants.
\end{itemize}

The outline of the paper is as follows: In Section \ref{sec:methods} we present our approach in detail (\ref{sec:system}), explain the  Differential Evolution algorithm (\ref{sec:diff_evolution}), and outline the evaluation study (\ref{sec:evaluation}). \mbox{Section \ref{sec:results}} contains the experimental results, which we discuss in \mbox{Section \ref{sec:discussion}}. The article is concluded by \mbox{Section \ref{sec:conclusion}}.

\section{Materials and Methods}
\label{sec:methods}

\subsection{System Design}
\label{sec:system}

\begin{figure}[ht]
    \centering
    \includegraphics[width=\textwidth]{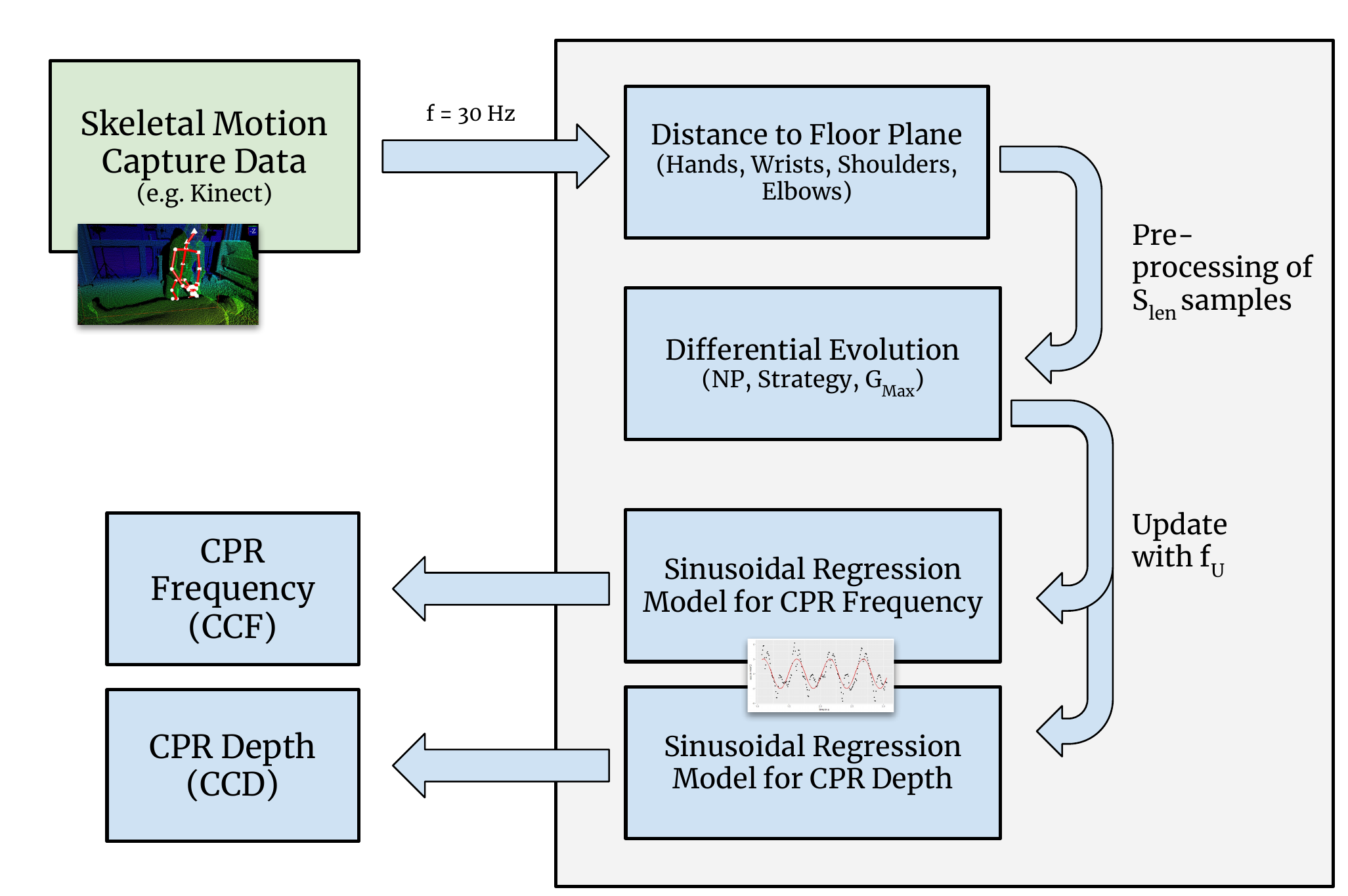}
    \caption{Structure of the RESKIN CPR parameter detection system}
    \label{fig:system_concept}
\end{figure}

\begin{figure}[ht]
    \centering
    \fbox{
    \includegraphics[width=0.85\textwidth,trim=0 0 0 1cm,clip]{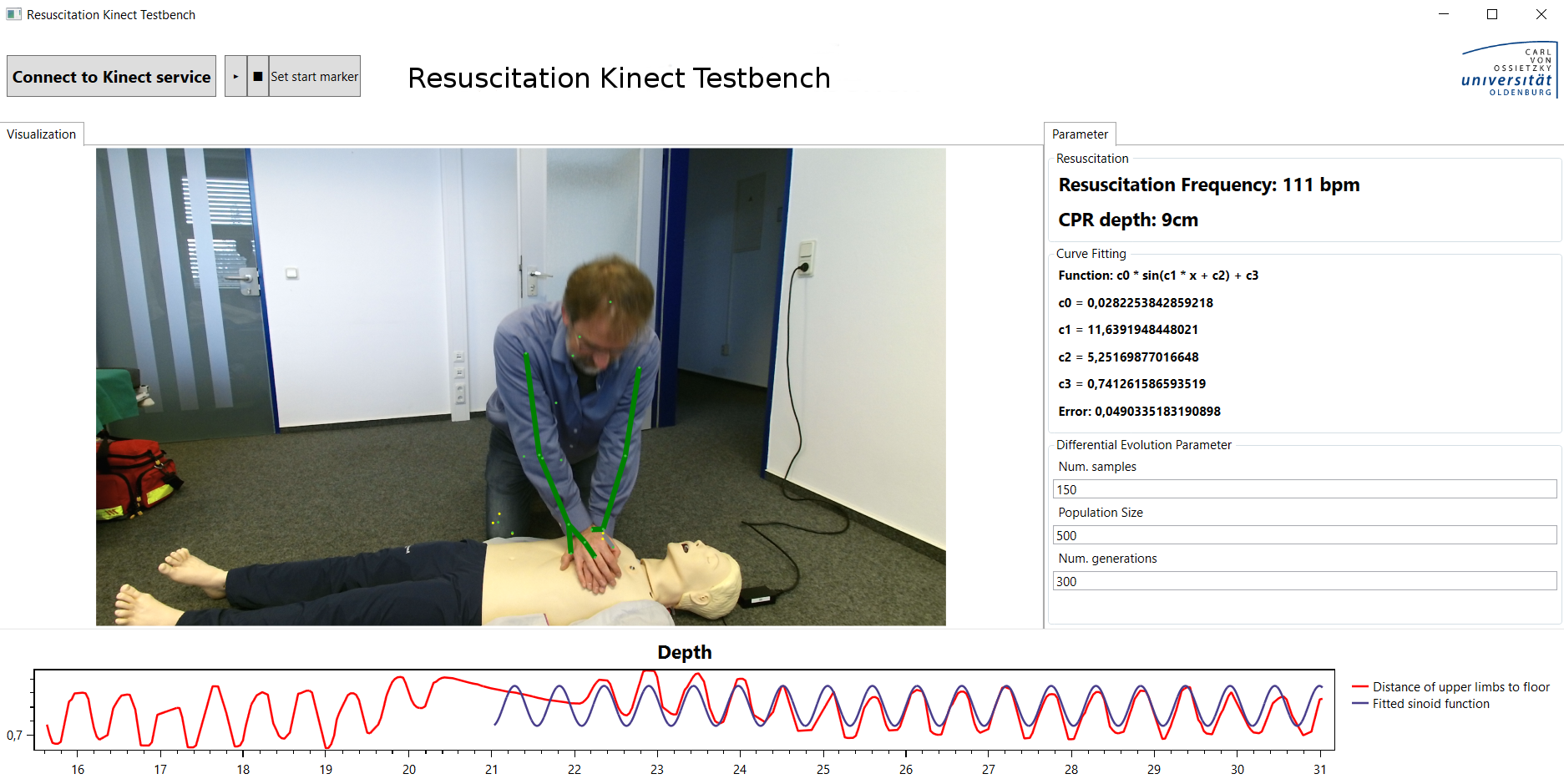}}
    \caption{Our software RESKIN processes Kinect motion data from a CPR training session. The graph at the bottom shows the distances between the floor and the limbs (red) and the fitted sinusoidal curve (blue). The Figure shows a typical screenshot of the application while processing motion data from the Kinect sensor. 
The RGB image is shown on the screen together with a green line overlay representing the arms. Additionally detected skeleton joints are marked as small yellow dots.}
    \label{fig:reskin}
\end{figure}

\noindent
An optical MoCap system can observe the full spatial spectrum, i.e. a 3-dimensional representation of the scenery. In our approach (see Figure \ref{fig:system_concept}), further referred to as \emph{RESKIN} (RESuscitation KINect), an optical MoCap system observes the scenery with a resuscitation task in progress. The trainee sits at the side of a mannequin (a simple sensor-free plastic mannequin or another compressible object can be used) and performs thorax compressions (see Figure \ref{fig:reskin}).  

The MoCap system (here: Microsoft Kinect v2) derives a skeleton from the optical data. RESKIN uses the joint positions of hands, wrists, elbows, or shoulders and therefore does not need a full skeleton, but at least an upper body. Artifacts in the area of the lower body (e.g. knees) can thus be safely ignored.

The Kinect sensor API notifies the application when a new data frame is available from the sensor (event-driven with \SI{30}{Hz}) and provides the application with a skeleton frame containing the joint positions and the floor plane estimation.
The application calculates the distance between the floor plane and the upper limb joints to effectively reduce the complexity of 3D motions (see Equation \ref{eq:dt}). On every ${f_{U}^{-1}}$ seconds the samples of the last $S_{len}$ seconds are used to update the sinusoid regression model (see Section \ref{sec:diff_evolution} for details). The DE algorithm is used to continuously adjust the models of the system. The hyperparameters $NP$ and $G_{max}$ of the DE algorithm are optimized once for this domain so that it can be executed with highest efficiency.

\subsection{Differential Evolution for Fitting Sinusoidal Curves}
\label{sec:diff_evolution}
\noindent
The MoCap system provides the spatial coordinates $v$ of the joints of the skeleton and a plane equation for the floor. Using this data, we continuously calculate the plane point distances for the sinusoidal fitting. Always two spatial coordinates (left $\vec v_{L}$ and right $\vec v_{R}$ body segment) are combined to one (Equation \ref{eq:combine}). Therefore there is one distance value $d(t)$ per time $t$.

The general equation of the floor plane \cite{blaschke1954analytische,weisstein2018plane} can be derived from the Kinect sensor's floor plane estimation that provides $\vec n = (n_x, n_y, n_z)$ and $a$:
\begin{align}
       n_x \cdot x_x + n_y \cdot x_y + n_z \cdot x_z - a &= 0
\end{align}

\noindent
Every measurement consists of a position vector for both left ($\vec v_L$) and right ($\vec v_R$) limbs (i.e. hands, elbows, shoulders, or wrist) that are combined to one vector $\vec v(t)$:

\begin{equation}
\label{eq:combine}
    \vec v(t) = \frac{\vec v_{L}(t) + \vec v_{R}(t)}{2}
\end{equation}

\noindent
Be $v_x(t), v_y(t), v_z(t)$ the three components of the vector $\vec v(t)$. Then the distance $d(t)$ between point vector and floor plane can be calculated with Equation \ref{eq:dt}, $d_t = d(t), d_{t} \in \mathbb{R}$:

\begin{equation}
\label{eq:dt}
    d(t) = \frac{n_x \cdot v_x(t) + n_y \cdot v_y(t) + n_z \cdot v_z(t) - a}{\sqrt{(n_x)^2 + (n_y)^2 + (n_z)^2}}
\end{equation}

\noindent
A set of this samples $d(t)$ forms a window $S$:
\begin{equation}
\label{eq:S}
S = \{d(t), \dots, d(t - S_{len})\}
\end{equation}

\noindent
The optimal value of $S_{len}$ is to be determined.

\subsubsection{Minimization Problem}
\begin{figure}[ht]
    \centering
    \includegraphics[width=\linewidth]{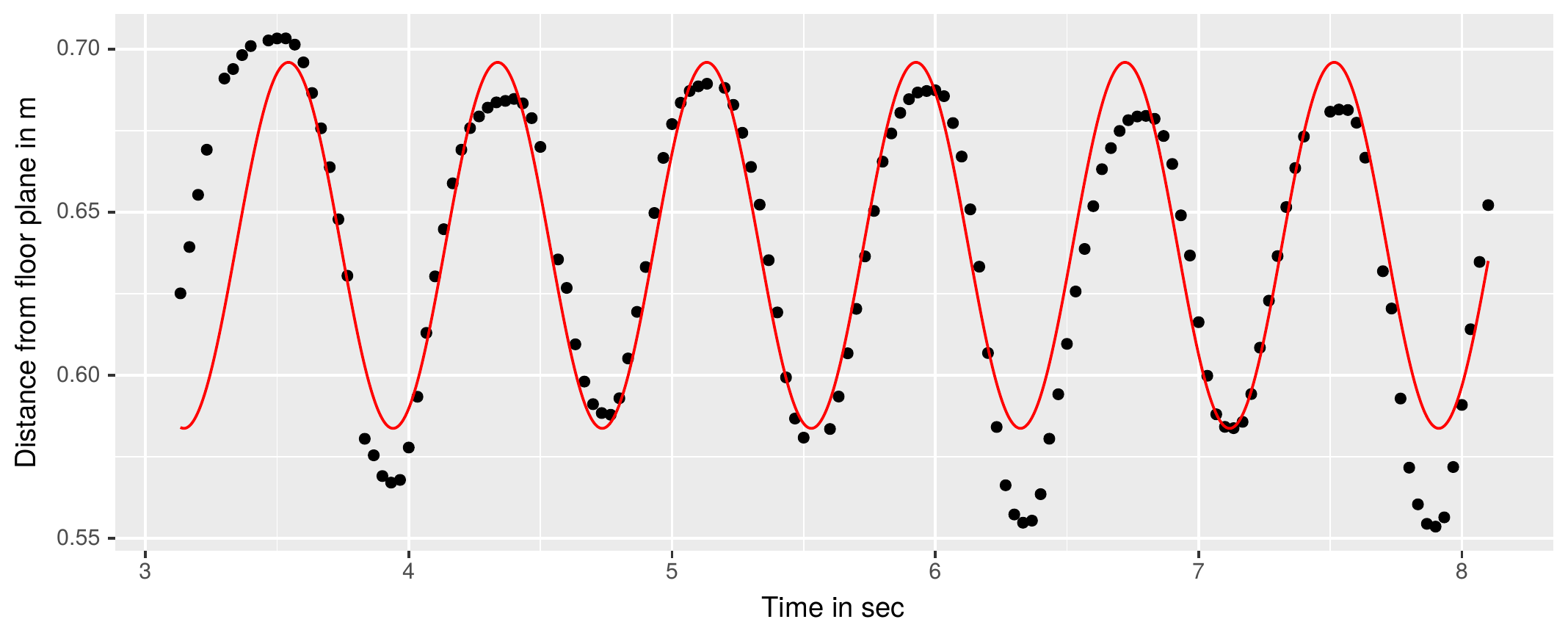}
    \caption{Fitting a sinusoid to Kinect distance data. In this excerpt from actual evaluation data, a data segment of $S_{len} = 5$ seconds was used.}
    \label{fig:sampleplot}
\end{figure}

\noindent
Our approach utilizes the periodic nature of the CPR to fit the time series of distances (upper limbs to ground) to a sine curve (see Figure \ref{fig:sampleplot}).

\noindent
The generic sine function with four parameters can be written as follows:
\begin{equation}
\label{eq:sinus}
    y(t) = A \cdot sin(2ft+\varrho)+D
\end{equation}

\noindent
As parameters, the amplitude $A$ and the frequency $f$ are of primary interest here since representing CCD and CCF. When assuming that the arms of a person performing CPR are rigid and exert orthogonal pressure on the patient's chest, then the relative distances of its arms between up and down movement are equal to the chest compression depth. Moreover, the frequency of low to high to low compression depth represents one compression cycle. To find a robust CPR frequency and compression depth, we fit the time-series data of upper-limb distances to a sinusoidal curve.
On a fitted function, parameter $f$ is the CPR frequency and $A$ is the compression depth. Parameter $D$ shifts the function values in Y-direction and $\varrho$ represents the phase shift.

\noindent
To fit the function $y(t)$, we minimize the cost function $c$ (here: RMSE) using an evolutionary approach. 
We can formulate this as a minimization problem:

\begin{equation} \label{eq:minproblem}
    \text{min } c = \sqrt{\frac{1}{|S|}\sum_{t=0}^{|S|} (S(t) - y(t))^2}
\end{equation}

\noindent
where $S$ is a vector (window) containing the joint-floor-distances $d(t)$ of the last $S_{len}$ seconds (see Equation \ref{eq:dt}). 

\subsubsection{Differential Evolution}
\noindent
The Differential Evolution algorithm \cite{Storn1997} is a generic evolutionary optimization algorithm that works particularly well with nonlinear, i.e., sinusoidal cost functions. DE searches and evaluates a parameter space concurrently and finds multiple near-optimal but distinct solutions to a problem.

As all evolutionary algorithms, DE is population-based and optimizes the individuals of the population (sometimes called agents) throughout several generations:

\begin{equation}
    x_{i,G} \text{ with } i =1..NP, G=1..G_{max}
\end{equation}

\noindent
Here $x_{i,G}$ is a 4-dimensional vector (because of the four unknowns of Equation \ref{eq:sinus}) of individual $i$ for generation $G$ representing one possible solution to our problem (see Equation \ref{eq:minproblem}).
So in every generation, $NP$ individuals are optimized up to $G_{max}$ generations.

The optimization is done between a transition from one generation $G$ to another generation $G+1$. Most evolutionary algorithms -- as does DE -- comprise the steps mutation, crossover, and selection, which are discussed in the following subsections. Each of these steps influences the convergence and runtime characteristics of the algorithm.

\paragraph{Mutation}

For every generation, a mutation step is performed for every individual $x_{i,G}$.
We used the following step with the fixed amplification factor $F=0.8$ \cite{Storn1997} :

\begin{equation}
    \label{eq:mut_str}
    v_{i,G+1} = x_{r_1,G} + F \cdot (x_{r_{2,G}} - x_{r_{3,G}})
\end{equation}


with $v$ the mutated individual and $r_1,r_2,r_3 \in \{1,2,\dots,NP \}, r_1 \neq r_2 \neq r_3 \neq i$  randomly chosen.
The strategy in Equation \ref{eq:mut_str} is labeled as \texttt{DE/rand/1/bin}, which is the original mutation strategy \cite{Storn1997}. 

\paragraph{Crossover}

The crossover step decides which of the four parameters of one individual are preserved in the next generation. For every parameter a uniform random number $r \in [0,1]$ is chosen. If $r \leq CR$ then the parameter from the mutant is chosen, otherwise the one from the original individual. Here, $CR$ is a constant value.

\paragraph{Selection}

The selection step decides which individual is passed to the next generation by evaluating it against the cost function (derived from the minimization problem, see Equation \ref{eq:minproblem}). In our approach the RMSE is summed up for every solution candidate $x_{i,G}$:

\begin{equation}
\label{eq:costfunc}
   c_{x_{i,G}} = \sum_{\tau=0}^{T} (S(\tau) - y_{x_{i,G}}(\tau))^2 
\end{equation}

with $S$ being a $T$-length vector of samples (joint-floor-distances) and $y$ the parameterized sinusoid function (Equation \ref{eq:sinus}) of individual $x_{i,G}$.
If $c_{v_{i,G+1}} < c_{x_{i,G}}$ the mutated individual $v_i$ is passed to the next generation otherwise $x_i$.

\subsection{Study Design}
\label{sec:evaluation}

\begin{figure}[ht]
    \centering
    \includegraphics[width=\textwidth]{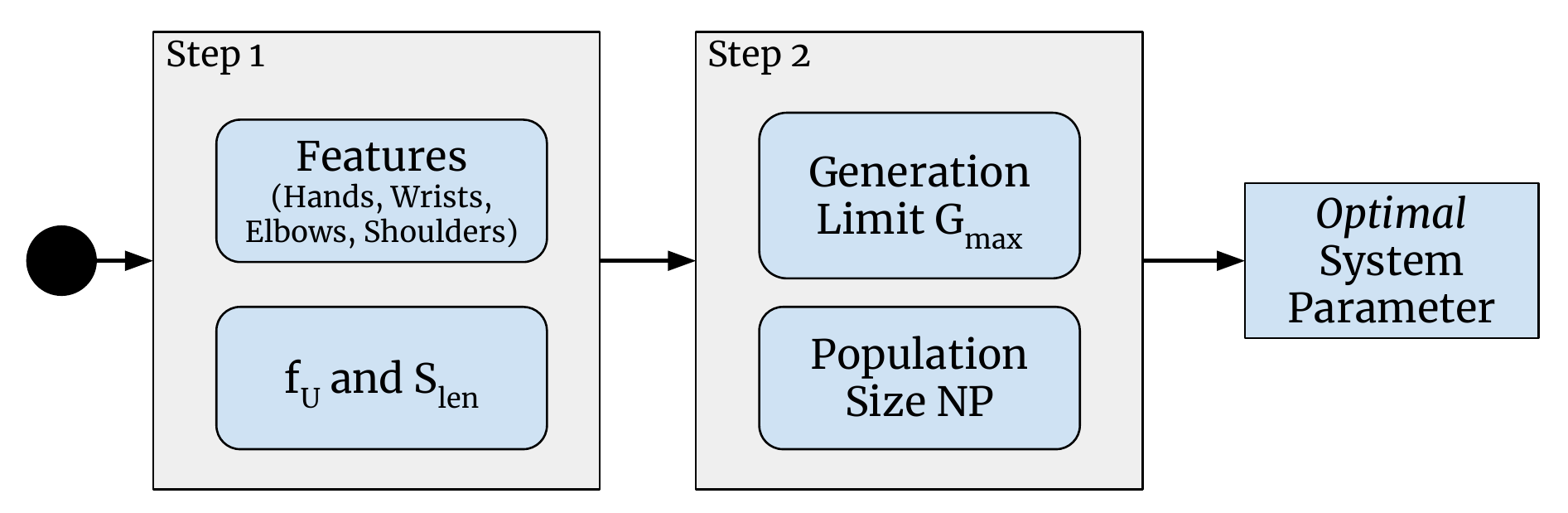}
    \caption{Study calculation steps overview.}
    \label{fig:eval_steps}
\end{figure}

\noindent
We have previously discussed the general suitability of the application of Differential Evolution algorithm to fit a sine curve describing the CPR parameters compression frequency and compression depth \cite{lins2018_ai4health18}.
The evaluation here comprises two further steps (see Figure \ref{fig:eval_steps}): 
\begin{enumerate}
    \item We investigated which of the upper limbs/joints, i.e. wrists, hands, elbows, or shoulders, are most suitable for the determination of CPR parameters. Additionally, we examined optimal values for model update frequency $f_U$ and the ideal window length $S_{len}$.
    \item We optimized the DE hyperparameters $NP$ and $G_{max}$ to improve the convergence specific to the domain further, thus minimizing the required computing power to retrieve a satisfactory result (Step 2, see Figure \ref{fig:eval_steps}).
\end{enumerate}

\subsubsection{Experimental setup}

\noindent
A Laerdal Resusci Anne Simulator mannequin was placed as reference system on the floor. Within the mannequin, sensors measure the depth of thorax compression and decompression, the frequency of the compressions and the volume of ventilation. The reference values were obtained from the tablet PC after the CPR training process was finished.

\begin{figure}[t]
    \centering
    \includegraphics[width=0.8\textwidth]{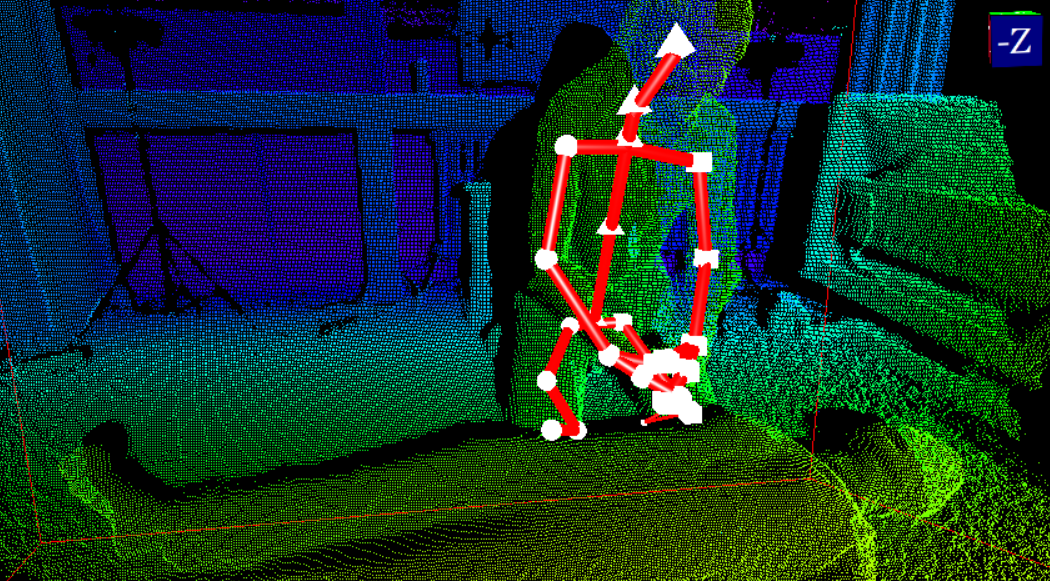}
    \caption{Voxel view of the 3D scene captured by the Kinect sensor with tracked skeleton as overview.}
    \label{fig:subject_3d}
\end{figure}

In addition, a Kinect sensor was placed at an approximate distance from the mannequin of \SI{1.5}{m} and at the height of about \SI{1}{m}. The person performing the CPR was placed on the other side of the mannequin facing the Kinect camera (see Figure \ref{fig:subject_3d}) although no precise position or posture nor precise viewing angle was enforced. The participant was asked to perform CPR compressions on the mannequin with standard CPR frequency and depth for about two minutes. The Kinect, as well as the mannequin, were collecting data, which was synchronized manually after the recording using the RGB color image of the Kinect. 


\subsubsection{Software and Parameter Settings}
\noindent
We performed two slightly different optimization steps that are described in this section. For the first step the R implementation of the DE algorithm (package DEoptim/2.2-4) \cite{Peterson2016, Ardia2014, ardia2015_deoptim} running on R/3.4.4 was used. For the second step the Python implementation of DE from the scipy.optimize (Version 0.19.0) package was used. 

\paragraph{General parameter settings of the cost function}

The sinus function $y(t) = A \cdot sin(2ft+\varrho)+D$ within the cost function (see Equation \ref{eq:costfunc}) to be minimized here has four unknown variables ($A, f, \varrho, D$), but the solutions to the function can be limited due to the specifics of our problem and the nature of the sinusoid itself. $A$ is the amplitude of the function. Valid values for $A$ are limited by the observed and derived Y-axis coordinates by the Kinect (assuming Y-axis pointing upwards, so $A \in (-2m, 2m)$). Same applies to $D$, the parameter that shifts the $y$ value of the function in a positive or negative direction, so we limit $D \in (-2m, 2m)$ as well. The parameter $\varrho$ is the phase shift of the function, which is limited by the timestamps of our samples, so $\varrho \in (S_{start}, S_{end})$. Finally, $f$ is the frequency of the sinusoid, which we can limit to a reasonable range \mbox{(60 cpm, 160 cpm)}, so $f\in (2\pi, \frac{16}{3}\pi)$.

The Kinect samples with \SI{30}{Hz}, so according to Nyquist-Shannon sampling theorem, the maximum reconstructible CPR frequency is 1800 cpm. The window length limits the minimum reconstructible CPR frequency. A window of \SI{5}{s} allows reconstruction of \SI{12}{cpm} (or \SI{6}{cpm}) when assuming a half period is enough to reconstruct the sinus curve. 

\paragraph{Step 1: System Model Parameter Optimization}
We vary the system parameters a.) window length $S_{len}$ and b.) update frequency $f_U$ used for the curve fitting (see Table \ref{tab:system_param_1}).

\begin{table}[h]
    \centering
        \caption{System parameter ranges for the first evaluation run.}
    \label{tab:system_param_1}
\begin{tabular}{c|c|c}
Parameter & $S_{len}$ & Update Frequency $f_U$\\\hline
Range & $\{1,2,3,4,5\}$ s & $\{0.25, 0.5, 0.75, 1.0, 1.5, 2.0\} s^{-1}$ \\
\end{tabular}
\end{table}

For every trial, we run the DE algorithm with the
parameters specified in Table \ref{tab:de_param_1}. The used hyperparameters are common defaults for many DE implementations. 
The number of individuals $NP$ is chosen by the rule of thumb of about "ten times the dimension" \cite{Storn1996}. The optimal value $NP$ is determined in the second optimization step.

\begin{table}[h]
    \centering
        \caption{DE hyperparameter for the first evaluation run.}
    \label{tab:de_param_1}
\begin{tabular}{c|c|c|c|c|c}
Parameter & CR & F & NP & $G_{max}$ & VTR \\\hline
Range & 0.5 & 0.8 & 50 & 500 & $0.0001$ \\
\end{tabular}
\end{table}

\paragraph{Step 2: Differential Evolution Hyperparameter Optimization}
Based on the results of the first run (see Section \ref{sec:mocapfitting}), $f_U = 1.0s^{-1}$ and $S_{len} = 3.0s$ were chosen as fixed values.
The number of individuals (population size) $NP$  influences how fast the DE algorithm is converging towards a good solution, i.e. if the algorithm converges fast, it requires fewer generations/iterations and therefore runs faster. 
Thus, we investigated combinations of $NP$ and $G_{max}$ within the given range $[10, 100]$. The hyperparameter used for this second optimization step are summarized in Table \ref{tab:de_param_2}.

\begin{table}[h]
    \centering
        \caption{DE parameter and parameter ranges for the second evaluation run.}
    \label{tab:de_param_2}
\begin{tabular}{c|c|c|c|c|c}
Parameter & CR & F & NP & $G_{max}$ \\\hline
Range & 0.5 & 0.8 & $\{10, 20, ..., 80, 90, 100, 150, 200\}$ & $\{10, 20, ..., 90, 100\}$ \\
\end{tabular} 

\end{table}
\noindent
The results of the DE hyperparameter optimization can be found in Table \ref{tab:result_hyper}.

\subsection{Comparing models with reference}

\begin{figure}
    \centering
    \includegraphics[width=\textwidth]{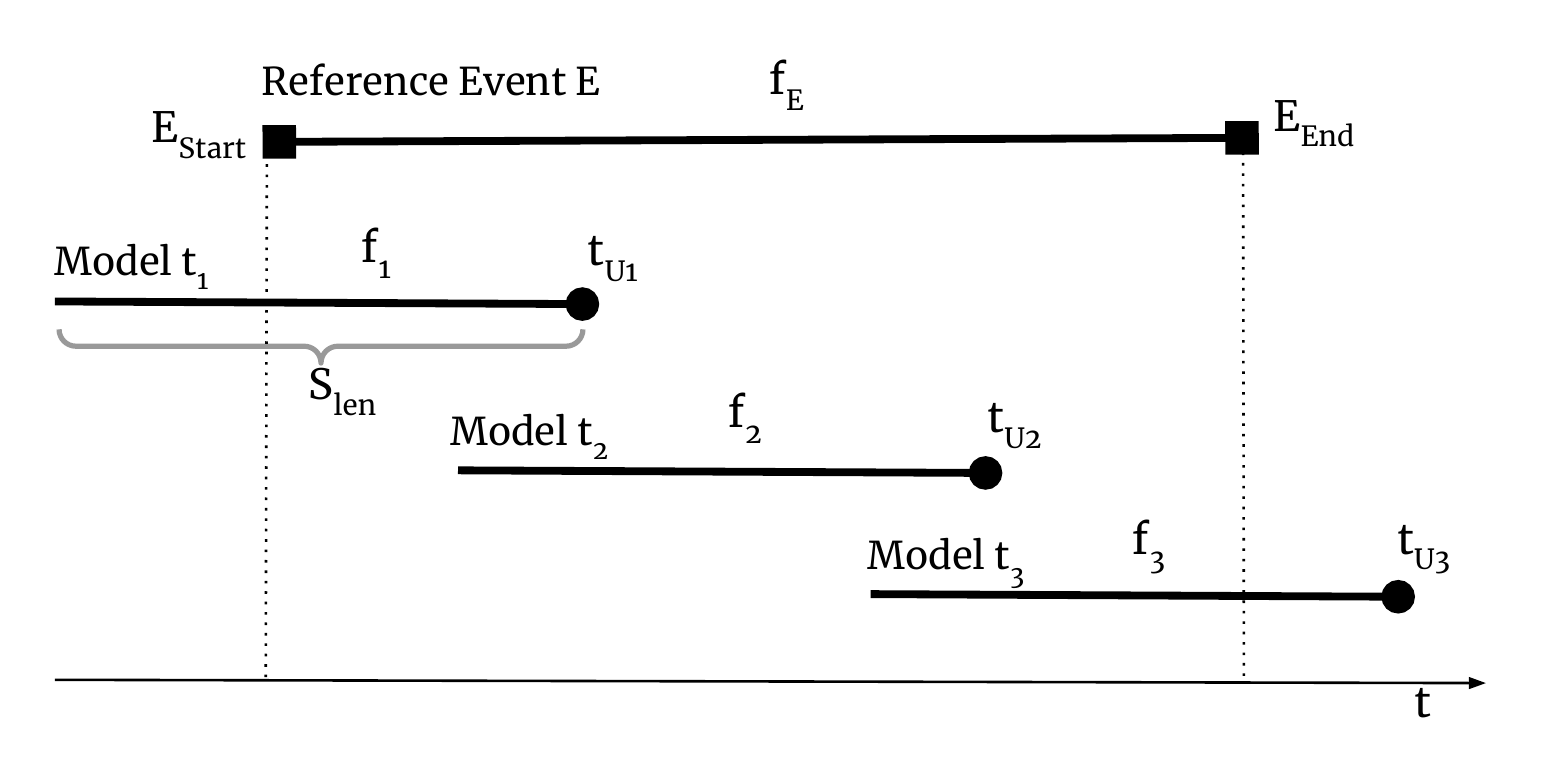}
    \caption{Overview of the different CPR frequencies to compare.}
    \label{fig:eval_calc}
\end{figure}

\noindent
The Resusci Anne mannequin records the CPR event-based, i.e. each compression cycle (compression and decompression) is considered one event. For every event, the start and end timestamps as well as the maximum compression depth and the current compression frequency are logged.

The RESKIN system adapts its current regression model every $f_U^{-1}$ seconds using the samples of the last $S_{len}$ seconds (see Equation \ref{eq:S}) beginning from the update timestamp $t_U$.
However, it is possible that $S_{len} \leq f^{-1}_{U}$, so the used datasets $S$ are interleaved (see Figure \ref{fig:eval_calc}). 
Also, one Resusci Anne event $E$ may be smaller, equal, or larger than $f^{-1}_{U}$ so that we must combine one or more model predictions before comparing it with $E$ (Equation \ref{eq:pred_combine}).
Thereby, the weighted mean of $n$ subsequent model predictions within each interval $(E_{Start}, E_{End})$ is calculated with the overlap ratio $\sigma$ representing the weight:

\begin{equation}
\label{eq:pred_combine}
    p(t) = \frac{1}{\sum_{i=1}^{n} \sigma_i} \sum_{i=1}^{n} \sigma_i f_i
\end{equation}



\FloatBarrier

\section{Results}
\label{sec:results}
\noindent
In this section, the optimized parameter settings and the results of the evaluation-study for the RESKIN system are presented. Of the recorded 35 study-participants, 7 had to be excluded from further consideration due to recording errors (malfunctions of storage system, Kinect, or Resusci Anne). An overall of 28 participants (21 female, 7 male, age 19-47 (median 24)) contributed 45 test runs of CPR. Four of the participants reported that they are well-trained in CPR. The participants were students and members of the University of Oldenburg. There were no inclusion or exclusion criteria.

\subsection{Model Prediction Accuracy}
\label{sec:mocapfitting}
\noindent
First, the influence of the window length $S_{len}$ of the motion capture recording and the update frequency $f_U$ on the accuracy of the model prediction was investigated. The model prediction was compared to the values of the reference system. Tables \ref{tab:error_shoulders}-\ref{tab:error_wrists} show the results of this comparison for each of the four different joints whose positions were used for the fitting. This clearly shows that the movements of the shoulder joints are best suited for adjusting the sinus model, followed by the elbow joints. The joints of the hands and wrists have higher errors compared to shoulders and elbows.

\begin{table}[h!]
    \centering
    
    \begin{minipage}{.5\textwidth}
        \caption{Median Absolute Error (MAE) (in cpm) for\\ CPR \textbf{frequency} prediction using \emph{shoulder} joints}
    \label{tab:error_shoulders}
    \begin{tabular}{c|c|c|c|c|c}
        \backslashbox{$f_U$}{$S_{len}$} &  \SI{1}{s} & \SI{2}{s} & \SI{3}{s} & \SI{4}{s} & \SI{5}{s}\\\hline
         \SI{0.25}{s^{-1}} & 2.12 & 1.54 & 1.44 & \textbf{1.40} & 1.43 \\\hline
         \SI{0.5}{s^{-1}} & 2.40 & 1.58 & 1.44 & 1.42 & 1.44\\\hline
         \SI{0.75}{s^{-1}} & 2.87 & 1.59 & 1.44 & 1.41 & 1.43\\\hline
         \SI{1.0}{s^{-1}} & 3.06 & 1.60 & 1.44 & 1.42 & 1.44\\\hline
         \SI{1.5}{s^{-1}} & 3.15 & 1.68 & 1.47 & 1.43 & 1.45\\\hline
         \SI{2.0}{s^{-1}} & 3.15 & 1.72 & 1.48 & 1.45 & 1.43\\
    \end{tabular}
    \hspace{1cm}
    \end{minipage}%
    \begin{minipage}{.5\textwidth}
    \caption{Median Absolute Error (in cpm) for CPR \textbf{frequency} prediction using \emph{elbows} joints}
    \label{tab:error_elbows}
    \begin{tabular}{c|c|c|c|c|c}
        \backslashbox{$f_U$}{$S_{len}$} &  \SI{1}{s} & \SI{2}{s} & \SI{3}{s} & \SI{4}{s} & \SI{5}{s}\\\hline
         \SI{0.25}{s^{-1}} & 3.01 & 1.82 & 1.55 & 1.50 & {1.49}\\\hline
         \SI{0.5}{s^{-1}} & 3.43 & 1.88 & 1.59 & {1.49} & 1.50\\\hline
         \SI{0.75}{s^{-1}} & 4.29 & 1.93 & 1.57 & 1.50 & 1.50\\\hline
         \SI{1.0}{s^{-1}} & 4.71 & 1.98 & 1.60 & 1.50 & {1.49}\\\hline
         \SI{1.5}{s^{-1}} & 5.01 & 2.15 & 1.62 & 1.50 & \textbf{1.48}\\\hline
         \SI{2.0}{s^{-1}} & 5.07 & 2.27 & 1.64 & 1.54 & {1.49}\\
    \end{tabular}
    \end{minipage}
    \end{table}

\begin{table}[h!]
    \centering
    
    \begin{minipage}{.5\textwidth}
    \caption{Median Absolute Error (in cpm) for CPR\\ \textbf{frequency} prediction using \emph{hand} joints}
    \label{tab:error_hands}
    \begin{tabular}{c|c|c|c|c|c}
        \backslashbox{$f_U$}{$S_{len}$} &  \SI{1}{s} & \SI{2}{s} & \SI{3}{s} & \SI{4}{s} & \SI{5}{s}\\\hline
         \SI{0.25}{s^{-1}} & 4.40 & 2.65 & 2.17 & 2.02 & 1.95\\\hline
         \SI{0.5}{s^{-1}} & 5.13 & 2.66 & 2.18 & 2.02 & 1.96\\\hline
         \SI{0.75}{s^{-1}} & 6.12 & 2.76 & 2.16 & 2.01 & 1.96\\\hline
         \SI{1.0}{s^{-1}} & 6.75 & 2.80 & 2.13 & 2.01 & 1.92\\\hline
         \SI{1.5}{s^{-1}} & 7.15 & 3.00 & 2.20 & 2.00 & 1.94\\\hline
         \SI{2.0}{s^{-1}} & 7.05 & 3.22 & 2.23 & 2.02 & \textbf{1.90}\\
    \end{tabular}
    \hspace{1cm}
    \end{minipage}%
    \begin{minipage}{.5\textwidth}
    \caption{Median Absolute Error (in cpm) for CPR \textbf{frequency} prediction using \emph{wrist} joints}
    \label{tab:error_wrists}
    \begin{tabular}{c|c|c|c|c|c}
        \backslashbox{$f_U$}{$S_{len}$} &  \SI{1}{s} & \SI{2}{s} & \SI{3}{s} & \SI{4}{s} & \SI{5}{s}\\\hline
         \SI{0.25}{s^{-1}} & 3.85 & 2.18 & 1.85 & 1.75 & 1.71\\\hline
         \SI{0.5}{s^{-1}} & 4.39 & 2.24 & 1.86 & 1.78 & 1.71\\\hline
         \SI{0.75}{s^{-1}} & 5.32 & 2.31 & 1.84 & 1.75 & 1.70\\\hline
         \SI{1.0}{s^{-1}} & 6.00 & 2.36 & 1.88 & 1.76 & 1.68\\\hline
         \SI{1.5}{s^{-1}} & 6.07 & 2.56 & 1.90 & 1.75 & 1.69\\\hline
         \SI{2.0}{s^{-1}} & 6.44 & 2.81 & 1.93 & 1.75 & \textbf{1.66}\\
    \end{tabular}
    \end{minipage}
    \end{table}
    
\FloatBarrier

While the prediction of the compression frequency by the models achieved only small errors (about $2\%$ error), the errors in the prediction of the compression depth are significantly higher. Again, the smallest errors are found when using the positions of the shoulder joints for the model fitting, followed by the elbow joints (see Tables \ref{tab:error_shoulder_depth}, \ref{tab:error_elbow_depth}, Tables for hands and wrists joints were omitted because the errors are only slightly different, if at all). In the optimum case, the error ranged from 1.0-1.5 cm (about $20\%$ error).

\begin{table}[h!]
    \centering
    \begin{minipage}{.5\textwidth}
    \caption{Median Absolute Error (in cm) for \emph{shoulder}\\ joints for compression \textbf{depth} prediction}
    \label{tab:error_shoulder_depth}
    \begin{tabular}{c|c|c|c|c|c}
        \backslashbox{$f_U$}{$S_{len}$} &  \SI{1}{s} & \SI{2}{s} & \SI{3}{s} & \SI{4}{s} & \SI{5}{s}\\\hline
         \SI{0.25}{s^{-1}} & 1.32 & 1.25 & 1.22 & 1.20 & 1.19\\\hline
         \SI{0.5}{s^{-1}} & 1.34 & 1.26 & 1.24 & 1.21 & 1.20\\\hline
         \SI{0.75}{s^{-1}} & 1.30 & 1.25 & 1.22 & 1.20 & 1.18\\\hline
         \SI{1.0}{s^{-1}} & 1.32 & 1.26 & 1.23 & 1.21 & 1.19\\\hline
         \SI{1.5}{s^{-1}} & 1.31 & 1.26 & 1.23 & 1.22 & 1.19\\\hline
         \SI{2.0}{s^{-1}} & 1.32 & 1.27 & 1.23 & 1.20 & 1.18\\
    \end{tabular}
    \hspace{1.5cm}
    \end{minipage}%
    \begin{minipage}{.5\textwidth}
    \caption{Median Absolute Error (in cm) for \emph{elbows} features for compression \textbf{depth} joints}
    \label{tab:error_elbow_depth}
    \begin{tabular}{c|c|c|c|c|c}
        \backslashbox{$f_U$}{$S_{len}$} &  \SI{1}{s} & \SI{2}{s} & \SI{3}{s} & \SI{4}{s} & \SI{5}{s}\\\hline
         \SI{0.25}{s^{-1}} & 1.35 & 1.36 & 1.34 & 1.29 & 1.25\\\hline
         \SI{0.5}{s^{-1}} & 1.35 & 1.36 & 1.34 & 1.29 & 1.26\\\hline
         \SI{0.75}{s^{-1}} & 1.35 & 1.36 & 1.34 & 1.29 & 1.26\\\hline
         \SI{1.0}{s^{-1}} & 1.36 & 1.36 & 1.35 & 1.32 & 1.29\\\hline
         \SI{1.5}{s^{-1}} & 1.40 & 1.37 & 1.36 & 1.32 & 1.28\\\hline
         \SI{2.0}{s^{-1}} & 1.39 & 1.37 & 1.35 & 1.33 & 1.29\\
    \end{tabular}
    \end{minipage}
    \end{table}

We discuss these results in Section \ref{sec:discussion} and also motivate the choice of $f_U = 1.0s^{-1}$ as the appropriate update frequency and window length $S_{len} = 3s$ for CCF prediction.


\FloatBarrier

\subsection{Sensitivity analysis}
\noindent
In order to determine the sensitivity of the system with respect to the parameters $f_U$ and $S_{len}$
a variance-based analysis is applied. Let $X_{f_U}$ represent the parameter $f_U$, $X_{S_{len}}$ the parameter $S_{len}$, $Y_{CCF}$ the random variable representing the CCF error, and $Y_{CCD}$ the CCD error. Then the \emph{correlation ratio} $ 0 \leq CR \leq 1 $ that describes the correlation between input parameter $X$ and output variables $Y$ can be defined as follows \cite{mckay1997nonparametric,Siebertz2017}:
\begin{equation}
    CR_{X} = \frac{Var[E(Y|X)]}{Var(Y)}
    \label{eq:correlation}
\end{equation}

The terms of Equation \ref{eq:correlation} can be calculated from the empirical results of the optimization. The expected value $E$ is estimated as the arithmetic mean of the error values \cite{mckay1997nonparametric,Arens2018}. The variance is determined directly. 
 As a result we get $CR_{X_{f_U} (CCF)} = 0.02$ and $CR_{X_{f_U} (CCD)} = 0.01 $ and $CR_{X_{S_{len}} (CCF|CCD)} \approx 1.0$ for the shoulder joints, if Equations \ref{eq:correlation} is applied to the results of the calculations in Section \ref{sec:mocapfitting}. This means that the model errors largely depend on the model parameter $S_{len}$ and the second parameter $f_U$ has only a small influence on the error.
 
\subsection{DE hyperparameter optimization}
\noindent
When optimizing with the Differential Evolution algorithm, the number of individuals used, i.e. the population size $NP$, as well as the (maximum) number of generations $G_{max}$ has a significant influence on how fast the algorithm converges to a global optimum. Therefore, in a second step, the influence of the two hyperparameters $NP$ and $G_{max}$ on the accuracy and convergence to a solution was investigated. Table \ref{tab:result_hyper}  shows how the error of our system changes compared to the reference system when the previously determined parameters of the system $f_U = 1.0s^{-1}$ and $S_{len} = 3.0$ are used in combination with varying hyperparameters.

\begin{landscape}
\begin{table}[]
    \centering
    \caption{Median frequency (cpm) / depth (cm) error between RESKIN and reference system with varying $NP$ and $G_{max}$.}
    \label{tab:result_hyper}
    \begin{tabular}{c|c|c|c|c|c|c|c|c|c|c}
        \diagbox{$NP$}{$G_{max}$} &  10 & 20 & 30 & 40 & 50 & 60 & 70 & 80 & 90 & 100\\\hline
        10 & 14.47 / 4.36 & 6.28 / 4.51 & 4.16 / 4.44 & 3.57 / 4.39 & 3.27 / 4.49 & 3.17 / 4.43 & 3.06 / 4.39 & 3.00 / 4.45 & 2.98 / 4.48 & 2.94 / 4.45 \\\hline
        20 & 13.02 / 4.49 & 5.32 / 4.42 & 3.80 / 4.43 & 3.31 / 4.51 & 3.14 / 4.23 & 3.09 / 4.51 & 2.98 / 4.52 & 2.97 / 4.54 & 2.94 / 4.41 & 2.95 / 4.42 \\\hline
        30 & 12.07 / 4.37 & 5.01 / 4.39 & 3.70 / 4.49 & 3.30 / 4.41 & 3.08 / 4.43 & 2.98 / 4.42 & 2.98 / 4.48 & 2.93 / 4.47 & 2.93 / 4.48 & 2.91 / 4.46 \\\hline
        40 & 11.34 / 4.34 & 4.64 / 4.47 & 3.55 / 4.49 & 3.20 / 4.49 & 3.03 / 4.46 & 2.99 / 4.46 & 2.96 / 4.46 & 2.97 / 4.45 & 2.93 / 4.39 & 2.93 / 4.39 \\\hline
        50 & 10.66 / 4.47 & 4.44 / 4.45 & 3.40 / 4.38 & 3.23 / 4.41 & 3.10 / 4.53 & 3.02 / 4.43 & 2.96 / 4.39 & 2.90 / 4.47 & 2.90 / 4.46 & 2.92 / 4.40 \\\hline
        60 & 10.48 / 4.43 & 4.24 / 4.49 & 3.41 / 4.42 & 3.17 / 4.46 & 3.06 / 4.44 & 2.95 / 4.43 & 2.94 / 4.51 & 2.98 / 4.45 & 2.96 / 4.40 & 2.92 / 4.44 \\\hline
        70 & 9.98 / 4.46 & 4.21 / 4.47 & 3.41 / 4.49 & 3.10 / 4.50 & 3.03 / 4.51 & 3.02 / 4.42 & 2.95 / 4.41 & 2.92 / 4.40 & 2.92 / 4.49 & 2.93 / 4.46 \\\hline
        80 & 9.43 / 4.34 & 4.14 / 4.44 & 3.31 / 4.51 & 3.13 / 4.49 & 3.01 / 4.55 & 2.98 / 4.34 & 2.97 / 4.47 & 2.95 / 4.46 & 2.93 / 4.43 & 2.90 / 4.39 \\\hline
        90 & 9.32 / 4.45 & 4.08 / 4.36 & 3.29 / 4.45 & 3.13 / 4.40 & 3.02 / 4.50 & 2.95 / 4.48 & 2.95 / 4.41 & 2.94 / 4.42 & 2.92 / 4.51 & 2.94 / 4.49 \\\hline
        100 & 9.03 / 4.49 & 4.00 / 4.45 & 3.37 / 4.40 & 3.11 / 4.45 & 3.01 / 4.44 & 2.96 / 4.52 & 2.94 / 4.45 & 2.92 / 4.43 & 2.94 / 4.47 & 2.94 / 4.42 \\\hline
        150 & 8.18 / 4.47 & 3.85 / 4.52 & 3.33 / 4.40 & 3.04 / 4.39 & 2.98 / 4.49 & 2.95 / 4.51 & 2.95 / 4.46 & 2.92 / 4.48 & 2.91 / 4.46 & 2.92 / 4.37 \\\hline
        200 & 7.88 / 4.42 & 3.66 / 4.47 & 3.14 / 4.50 & 3.08 / 4.41 & 3.00 / 4.43 & 2.93 / 4.51 & 2.93 / 4.41 & 2.95 / 4.35 & 2.91 / 4.46 & 2.92 / 4.40 \\
    \end{tabular}
\end{table}
\end{landscape}

The errors for the prediction of the CCF show a dependency of $NP$ and $G_{max}$ in the range $[10..200]$ and $[10..100]$ respectively. At $NP = 10$ and $G_{max} = 10$ the errors are expected to be highest and decrease with increasing $NP, G_{max}$. For both parameters the error is approximately logarithmically decreasing.
The error for the prediction of the compression depth (CCD) for all $NP,G_{max}$ is mostly constant at about \SI{4.4}{cm}. 
\FloatBarrier

\section{Discussion}
\label{sec:discussion}

\noindent

\noindent
The purpose of this study was to evaluate the accuracy of a new approach for deriving CPR quality parameters based on skeletal data from an optical MoCap system using the DE algorithm to fit sinusoidal models. It was evaluated against a Laerdal Resusci Anne training mannequin as gold standard which joint positions are best suited for fitting MoCap data to the model. To adjust the model for deriving the CPR frequency, the data of the shoulders result in the least error. In a previous examination \cite{lins2018_ai4health18}, the fitting with the elbows had the least error, but the error of the shoulder was only marginally higher. The shoulders are usually well visible from the sensors of the MoCap system, i.e. there are only a few errors in skeletal detection, which benefits the robust recognition of the CPR frequency.

In order to select an appropriate update frequency $f_U$ and window length $S_{len}$ based on the results, it must be taken into account that both parameters affect the performance and responsiveness of the approach. A high update frequency increases the responsiveness at the expense of performance. A large window length increases the computational effort, reduces the responsiveness, but stabilizes the results against outliers. Thus, we recommend $f_U = 1.0s^{-1}$ and $S_{len} = 3.0$ here.

This tradeoff must also include the parameters $NP, G_{max}$, which we examined in the second step. We were able to show that the error for the CCF prediction decreases logarithmically for linearly increasing $NP, G_{max}$. Therefore, there is an optimal range where the prediction error and computational effort remain within a practical  range. According to the results in Table \ref{tab:result_hyper} it can be seen that a $NP = 50$ and $G_{max} = 80$ is sufficient to achieve a satisfactory CCF prediction (Median Error $\pm2.90$ cpm). This makes it possible to detect whether a trainee reanimates in the recommended CCF range of $110 \pm 10$ cpm, while the computational effort remains comparatively low.

For the prediction of the compression depth CCD, the range $NP=200, G_{max}=100$ is not sufficient to achieve a satisfactory result. A convergence of the error is not recognizable in the examined range. Probably a much more accurate fitting of the sine model is necessary for CCD prediction, because with $G_{max} = 500$ a much smaller error between $1.18-1.32$ cm was achieved.

With a median error between $1.18$ cm, however, it is sufficient to make basic quality statements. An insufficient compression depth of, for example, \SI{3}{cm} can be reliably distinguished from an adequate compression depth of \SI{5.5}{cm}, so basic feedback to the trainee is possible. 
The reason for the comparatively high error rate in determining the compression depth is probably the inaccuracy of the Kinect MoCap system. In the literature, Kinect v2 deviations in the centimeter range (up to $\pm \SI{10}{cm}$ for some joints) are given for the accuracy of the skeleton tracking \cite{otte2016, Wang2015}. Therefore the values determined here -- despite the high errors rates -- are an excellent approximation to the optimum possible with this sensor. 

The accuracy with which our sinusoidal model can be used to derive the CPR frequency is in the range of other CPR training systems. We use a 3-second window to calculate the frequency with \SI{2.90}{cpm} median error. This window size seems to be fairly well suited for the frequency prediction, regardless of the technology used, as Ruiz de Gauna et al. \cite{gauna2016} for example use a 2-second window and report a median error of 5.9\% (about $\pm$ \SI{6.5}{cpm} with 110 cpm) for determining the CPR frequency with their IMU-based approach.

The approach of recording the motion data in a realistic setting shows the practicality of the model but does not exploit the maximum precision of the model. It has not been investigated how precisely the adjustment responds to sudden frequency changes, as this is not necessary for use in a CPR feedback system. Relevant for CPR training is the detection of continuous (unnoticeable to the user) changes in training parameters. This means that the feedback system does not give immediate feedback on a single outlier from the target range (e.g. frequency is too low), but only if the training parameters are outside the target range for a few seconds or if they show a negative trend.

In summary, our study shows that the position data of shoulder joints derived from optical MoCap systems can be used to fit sinusoidal models. With this model, robust CCF (CCD with limitations) quality parameters can be derived from realistic CPR training scenarios.

\section{Conclusion}
\label{sec:conclusion}
\noindent
We presented a software system that uses motion data from an RGB-D (Kinect) sensor and the Differential Evolution (DE) optimization algorithm to dynamically fit sinusoidal curves to derive frequency (CCF) and depth (CCD) parameters for CPR training.
We evaluated the approach with 45 trials of 28 different participants and tested the data of four different limb regions (hands, elbows, wrists, and shoulders) for their suitability to derive the parameters. Using the shoulder features of the Kinect skeleton, the results for the frequency determination show a median deviation of $\pm 1.44$ cpm (with $G_{max} = 500$) compared to the results of a Laerdal Resusci Mannequin (acting as the gold standard). We have also investigated which window length ($S_{len}$) and model update frequency ($f_U$) is the best tradeoff between adaptability to CPR frequency changes and precision. We found that a window size of $S_{len} = 3s$ and $f_U = 1.0s^{-1}$ reduces the error between model prediction and Resusci Anne to a reasonable minimum while preserving computing resources.

We investigated the dependence of the prediction error on the hyperparameters $NP$ and $G_{max}$ and were able to show a logarithmic decrease of the error for the parameter range of $[10..200]$ and $[10..100]$ respectively. With $NP = 50, G_{max} = 80$ a practical deviation of $\pm 2.90$ cpm could be determined. This makes it possible to detect whether a trainee reanimates in the recommended CCF range of $110 \pm 10$ cpm, while the computational effort remains comparatively low.

We summarize our findings and recommendations in Table \ref{tab:recommendations}.

\begin{table}[h]
    \centering
    \begin{tabular}{c||c|c|c|c|c||c}
      & Joints & $f_U$ & $S_{len}$ & $NP$ & $G_{max}$ & Median error\\\hline
    CCF & Shoulders & $1.0s^{-1}$ & \SI{3.0}{s} & 50 & 80 & $\pm 2.90$ cpm\\
    CCD & Shoulders & $2.0s^{-1}$ & \SI{5.0}{s} & 50 & 500 & $\pm 1.18$ cm
    \end{tabular}
    \caption{Recommended parameter for CCF and CCD predictions}
    \label{tab:recommendations}
\end{table}

In addition to CCF and CCD, which are indeed the essential quality parameters for CPR, the flexion of the arms during the compression process could be considered as well. With the data of the Kinect, which can derive the positions of the shoulder, elbow, hands, and wrists, a measure of the flexion of the compression could be recorded. It remains to be investigated if the differences in compression depth between shoulder and hand models can be used to derive a degree of arm flexion.
Also, in the current method, the fitting of the model is continuously adjusted, but the DE algorithm assumes an entirely new fitting. It could still be investigated how the convergence and performance of the algorithm change when parts of the population (i.e., the possible solutions) are included, thus allowing a faster adaptation of the model. The difficulty here is also to react to extreme changes in the trainee's CPR and not to prematurely exclude parts of the solution space.

In conclusion, the approach described in this paper shows comparable and practical results that can contribute to novel CPR training devices that are suitable for use in both the lab, clinic, and field.

\section*{Acknowledgements}
\noindent
We appreciate the support of Rainer R\"ohrig and Myriam Lipprandt in the development of the initial research idea.
This work was supported by the funding initiative Nieders\"achsisches Vorab of the Volkswagen Foundation and the Ministry of Science and Culture of the Lower Saxony State (MWK) as a part of the Interdisciplinary Research Centre on Critical Systems Engineering for Socio-Technical Systems II.
The optimizations were performed at the HPC Cluster CARL, located at the Carl von Ossietzky University of Oldenburg (Germany) and funded by the DFG through its Major Research Instrumentation Programme (INST 184/157-1 FUGG) and the MWK. 



\section*{References}

\bibliographystyle{elsarticle-num}
\bibliography{cprsinusoidal}



\end{document}